\documentclass[runningheads]{llncs}
\usepackage{graphicx}
\usepackage{comment}
\usepackage{amsmath,amssymb} 
\usepackage{color}

\usepackage{epsfig}
\usepackage{mathtools}
\usepackage{multirow}
\usepackage{algorithm}
\usepackage{algorithmic,eqparbox,array}
\usepackage{booktabs}
\usepackage{makecell}
\usepackage{verbatim} 
\usepackage{tikz}
\usepackage{pgfplotstable}

\newcommand{\etal}{\textit{et al}. }
\newcommand{\ie}{i.e., }
\newcommand{\eg}{eg., }
\def\checkmark{\tikz\fill[scale=0.4](0,.35) -- (.25,0) -- (1,.7) -- (.25,.15) -- cycle;}

\newcolumntype{C}{>{\centering\arraybackslash} m{1.5cm} }

\begin{document}
\pagestyle{headings}
\mainmatter

\title{Short-Term and Long-Term Context Aggregation Network for Video Inpainting} 


\titlerunning{Short-Term and Long-Term Context Aggregation Net for Video Inpainting}
%
\author{Ang Li\inst{1} \and
Shanshan Zhao\inst{2} \and
Xingjun Ma\inst{3} \and
Mingming Gong\inst{4} \and
Jianzhong Qi\inst{1} \and
Rui Zhang\inst{1} \and
Dacheng Tao\inst{2} \and
Ramamohanarao Kotagiri\inst{1} 
}
\authorrunning{Ang Li et al.}
%
\institute{School of Computing and Information Systems, The University of Melbourne, Australia 
\email{\{angl4@student., jianzhong.qi@, rui.zhang@, kotagiri@\}unimelb.edu.au} \and
UBTECH Sydney AI Centre, School of Computer Science, Faculty of Engineering, The University of Sydney, Darlington, NSW 2008, Australia
\email{\{szha4333@uni., dacheng.tao@\}sydney.edu.au} \and
School of Information Technology, Deakin University, Geelong, Australia
\email{daniel.ma@deakin.edu.au} \and
School of Mathematics and Statistics, The University of Melbourne, Australia
\email{mingming.gong@unimelb.edu.au}}
\maketitle

\begin{abstract}
Video inpainting aims to restore missing regions of a video and has many applications such as video editing and object removal.
However, existing methods either suffer from inaccurate short-term context aggregation or rarely explore long-term frame information.
In this work, we present a novel context aggregation network to effectively exploit both short-term and long-term frame information for video inpainting. 
In the encoding stage, we propose \textbf{boundary-aware short-term context aggregation}, which aligns and aggregates, from neighbor frames, local regions that are closely related to the boundary context of missing regions into the target frame\footnote{The target frame refers to the current input frame under inpainting.}.
Furthermore, we propose \textbf{dynamic long-term context aggregation} to globally refine the feature map generated in the encoding stage using long-term frame features, which are dynamically updated throughout the inpainting process.
Experiments show that it outperforms state-of-the-art methods with better inpainting results and fast inpainting speed.
\end{abstract}


\section{Introduction}\label{secIntro}
Video inpainting aims to restore missing regions in a video with plausible contents that are both spatially and temporally coherent \cite{huang2016temporally,newson2014video}.
It can benefit a wide range of practical video applications such as video editing, damage restoration, and undesired object removal.
Whilst significant progress has been made in image inpainting \cite{iizuka2017globally,liu2018image,pathak2016context,Xiong_2019_CVPR,Yang_2017_CVPR,yu2018generative,zeng2019learning}, it is challenging to extend image inpainting methods to solve the video inpainting problem.
Directly applying image inpainting methods on individual video frames may lose the inter-frame motion continuity and content dependency, which causes temporal inconsistencies and unexpected flickering artifacts.

Traditional video inpainting methods \cite{huang2016temporally,newson2014video,wexler2004space} utilize patch-based optimization strategies to fill missing regions with sampled patches from known regions.
These methods often suffer from limited effectiveness and vulnerability to complex motions.
Recently, deep learning-based video inpainting methods \cite{kim2019deep,lee2019copy,oh2019onion,wang2019video} have improved the inpainting performance by a large margin.
Most of them use encoder-decoder structures following a frame-by-frame inpainting pipeline to borrow information from reference frames\footnote{The reference frames refer to other frames from the same video.} and perform different types of context aggregation to restore the target frame.

\begin{figure}[t!]
    \centering
    \setlength{\tabcolsep}{0.1em}
    {

\begin{tabular}{ccccc}
        
\rotatebox[origin=l]{90}{\scriptsize Input}&
\includegraphics[width=2.2cm]{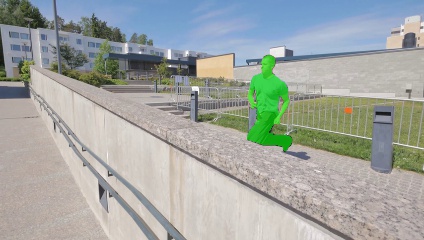}&
\includegraphics[width=2.2cm]{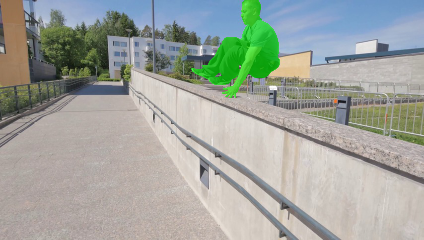}&
\includegraphics[width=2.2cm]{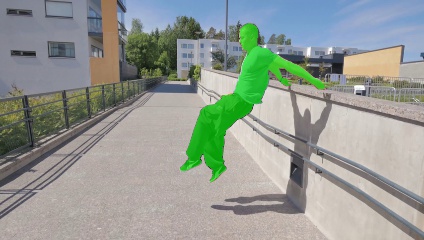}&
\includegraphics[width=2.2cm]{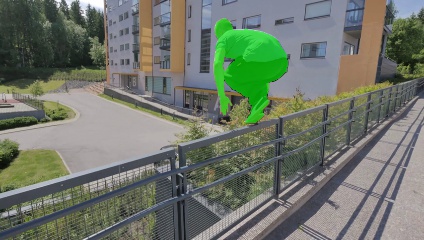}\\

\addlinespace[-0.1em]

\rotatebox[origin=l]{90}{\scriptsize CPNet}&
\includegraphics[width=2.2cm]{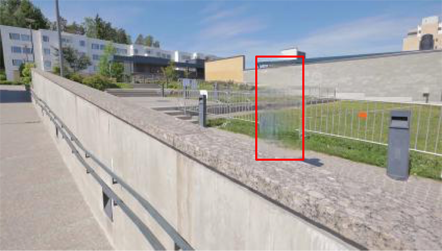}&
\includegraphics[width=2.2cm]{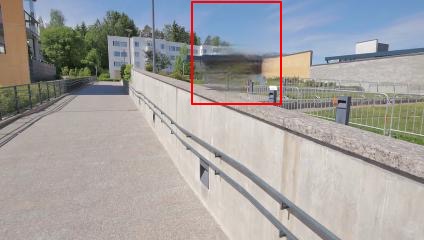}&
\includegraphics[width=2.2cm]{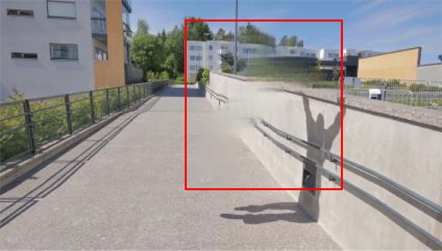}&
\includegraphics[width=2.2cm]{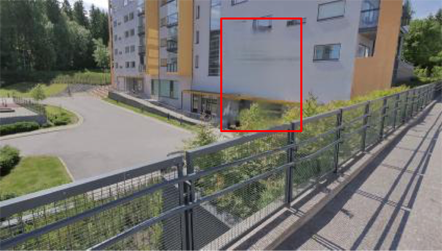}\\

\addlinespace[-0.1em]

\rotatebox[origin=l]{90}{\scriptsize FGNet}&
\includegraphics[width=2.2cm]{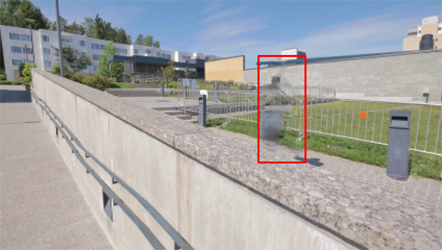}&
\includegraphics[width=2.2cm]{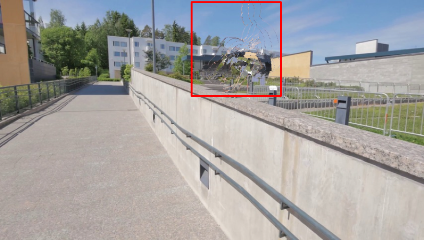}&
\includegraphics[width=2.2cm]{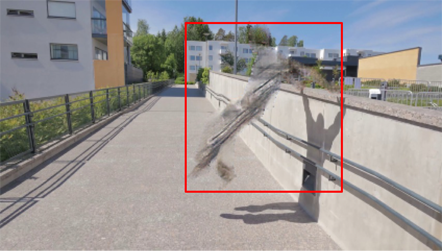}&
\includegraphics[width=2.2cm]{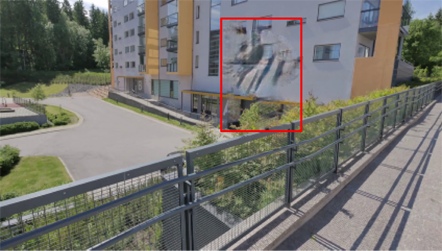}\\

\addlinespace[-0.1em]

\rotatebox[origin=l]{90}{\scriptsize Ours}&
\includegraphics[width=2.2cm]{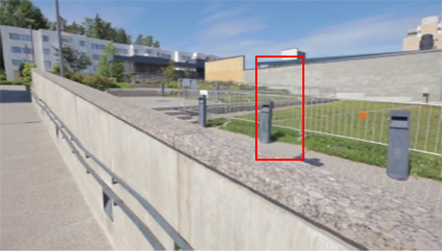}&
\includegraphics[width=2.2cm]{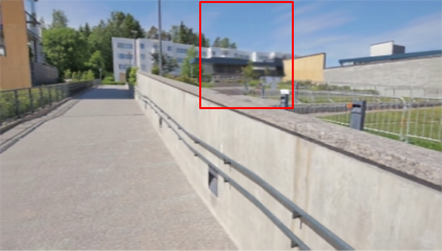}&
\includegraphics[width=2.2cm]{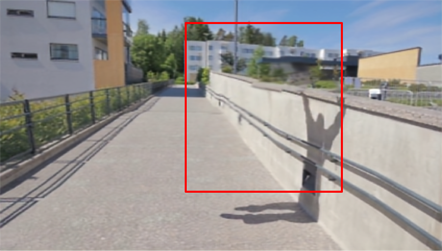}&
\includegraphics[width=2.2cm]{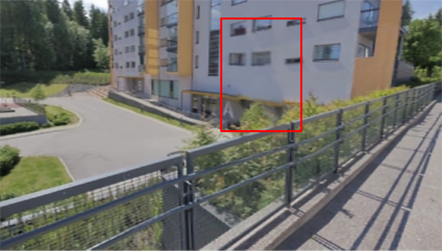}\\

\end{tabular}

}
   
\caption{Comparison with state-of-the-art CPNet \cite{lee2019copy} and FGNet \cite{xu2019deep}. The green areas in the input frames are the missing regions. Best viewed at zoom level 400\%.}
\label{QualitativeCompare}

\end{figure}

\begin{figure*}[t!]
\centering
\setlength{\tabcolsep}{0.1em}
{
\begin{tabular}{ccccc}

\includegraphics[width=2.2cm]{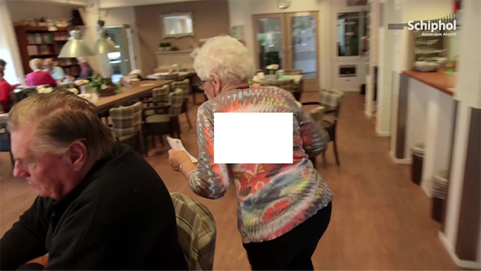}&
\includegraphics[width=2.2cm]{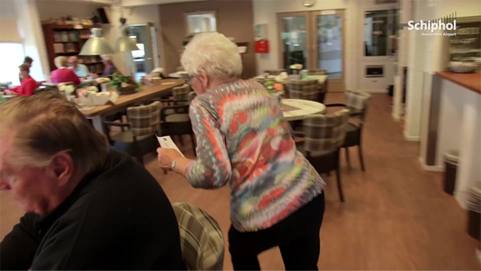}&
\includegraphics[width=2.2cm]{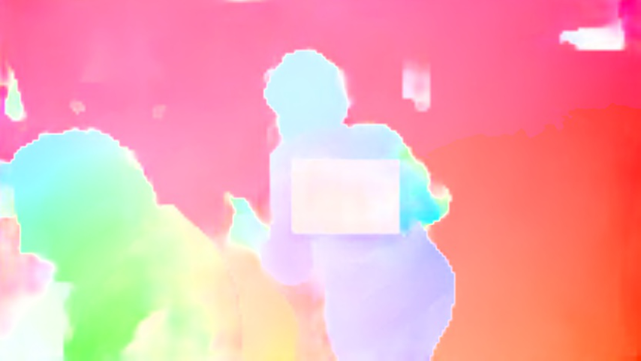}&
\includegraphics[width=2.2cm]{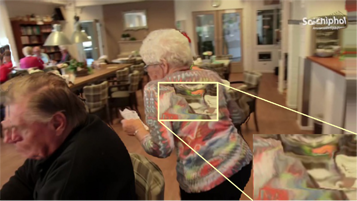}&
\includegraphics[width=2.2cm]{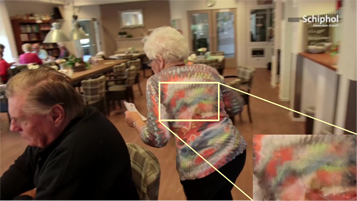}

\\

   target frame  & reference frame  & optical flow   & warped frame   & Ours
  \\
 $X_{t}$  & $X_{i}$  &  $F_{t \rightarrow i}$ & $W(F_{t \rightarrow i},X_{i})$ &
  \\

\end{tabular}
}

\caption{An example of missing regions' negative effects on flow-warping-based aggregation. Given a target frame $X_{t}$ and a reference frame $X_{i}$, we get the optical flow $F_{t \rightarrow i}$ between them using a pretrained flow estimator. Then we warp $X_{i}$ onto $X_{t}$ and get the warped frame $W(F_{t \rightarrow i},X_{i})$. Heavy distortions can be found in $F_{t \rightarrow i}$ and $W(F_{t \rightarrow i},X_{i})$ within the missing regions. 
Our network alleviates this problem, producing more accurate aggregation.}
\label{flow} 

\end{figure*}

In spite of the encouraging results, deep learning-based methods still need to overcome the following limitations. 
First, they fail to make effective usage of short-term and long-term reference information in the input video.
Studies \cite{kim2019deep,wang2019video} restrict the range of reference frames to nearby (short-term) frames of the target frame so as to maintain temporal consistency. 
When dealing with diverse motion patterns in videos (\eg slowly moving views or objects), short-term frames alone cannot provide sufficient information to restore the target frame.
Other methods \cite{lee2019copy,oh2019onion} often sample a set of fixed frames from the input video (\eg every 5-th frame) as the reference frames.
Although this can exploit some long-term information, it tends to include irrelevant contexts, reduce temporal consistency, and increase the computation time.
Second, how to achieve accurate context aggregation remains challenging.
Since missing regions contain no visual information, it is difficult to find the most related local regions in reference frames for accurate context aggregation.
For example, a recent method \cite{kim2019deep} uses estimated optical flows to warp reference frames onto the target frame and further aggregate them together.
As shown in Figure~\ref{flow}, the flow information within the missing region is inaccurate and is distorted.
This will cause unexpected artifacts when using the distorted flow to do warping and context aggregation, and the distortion artifacts will be propagated and accumulated during the encoding process.
While using more fixed reference frames from the input video may help mitigate the negative effects of missing regions, it inevitably brings more irrelevant or even noisy information into the target frame, which inadvertently does more harm than good.


In this paper, we aim to address the challenges above in a principled manner from the following three aspects: 
(1) We propose a novel framework for video inpainting that integrates the advantages of both short-term and long-term reference frames.
Different from existing methods that only conduct context aggregation at the decoding stage, we propose to start context aggregation at the encoding stage with short-term reference frames. This can help provide more temporally consistent local contexts.
Then, at the decoding stage, we refine the encoding-generated feature map with a further step of context aggregation on long-term reference frames. This refinement can deal with more complex motion patterns.
(2) To better exploit short-term information, we propose \emph{boundary-aware short-term context aggregation} at the encoding stage.
Different from existing methods, here, we pay more attention to the boundary context of missing regions.
Our intuition is that, in the target frame, the boundary area around the missing regions is more related to the missing regions than other areas of the frame.
Considering the spatial and motion continuity of videos, if we can accurately locate and align the boundary context of missing regions with the corresponding regions in the reference frames, it would improve both the spatial and temporal consistency of the generated contents.
This strategy can also alleviate the impact of missing regions at context aggregation.
(3) To better exploit long-term information, we propose a \emph{dynamic long-term context aggregation} at the decoding stage.
Since different videos have different motion patterns (\eg slow moving or back-and-forth moving), they have different contextual dependency between frames.
Therefore, it is necessary to eliminate frames that are largely irrelevant with the target frame.
Our dynamic strategy aims for the effective usage of long-term frame information.
Specifically, instead of simply using fixed reference frames, we propose to dynamically update the long-term reference frames used for inpainting, according to similarities of other frames to the current target frame.


In summary, our main contributions are:
\begin{itemize}
  \item We propose a novel framework for video inpainting that effectively integrates context information from both short-term and long-term reference frames.

  \item We propose a \textit{boundary-aware short-term context aggregation} to better exploit the context information from short-term reference frames, by using the boundary information of the missing regions in the target frames.
  
  \item We propose a \textit{dynamic long-term context aggregation} as a refinement operation to better exploit the context information from dynamically updated long-term reference frames.

  \item We empirically show that our proposed network outperforms the state-of-the-art methods with better inpainting results and fast inpainting speed.
\end{itemize}

\section{Related Work}\label{secBackground}
\subsection{Image Inpainting}
Traditional image inpainting methods \cite{barnes2009patchmatch,bertalmio2003simultaneous} mostly perform inpainting by finding pixels or patches outside missing regions or from the entire image database.
These methods often suffer from low generation quality, especially when dealing with complicated scenes or large missing regions \cite{iizuka2017globally,pathak2016context}.

Image inpaint methods \cite{iizuka2017globally,Ang_2019_IJCNN,Ang_2019_IJCAI,liu2018image,pathak2016context,Sagong_2019_CVPR,Xiong_2019_CVPR,Yang_2017_CVPR,yeh2017semantic,yu2018generative,zeng2019learning} with deep learning techniques have attracted wide attention.
Pathak \etal \cite{pathak2016context} introduce the \textit{Context Encoder} (CE) model where a convolutional encoder-decoder network is trained with the combination of an adversarial loss \cite{goodfellow2014generative} and a reconstruction loss.
Iizuka \etal \cite{iizuka2017globally} propose to utilize global and local discriminators to ensure consistency on both entirety and details.
Yu \etal \cite{yu2018generative} propose the contextual attention module to restore missing regions with similar patches from undamaged regions in deep feature space.

\subsection{Video Inpainting}
Apart from spatial consistency in every restored frame, video inpainting also needs to solve a more challenging problem: how to make use of information in the whole video frame sequence and maintain temporal consistency between the restored frames.
Traditional video inpainting methods adopt patch-based strategies.
Wexler \etal \cite{wexler2004space} regard video inpainting as a global optimization problem by alternating between patch search and reconstruction steps.
Newson \etal \cite{newson2014video} extend this and improve the search algorithm by developing a 3D version of PatchMatch \cite{barnes2009patchmatch}.
Huang \etal \cite{huang2016temporally} introduce the optical flow optimization in spatial patches to enforce temporal consistency.
These methods require heavy computations, which limit their efficiency and practical use.

Recently, several deep learning-based methods \cite{chang2019free,kim2019deep,lee2019copy,oh2019onion,wang2019video,xu2019deep,zhang2019internal} have been proposed.
These works can be divided into two groups.
The first group mainly relies on short-term reference information when inpainting the target frame.
For example, VINet \cite{kim2019deep} uses a recurrent encoder-decoder network to collect information from adjacent frames via flow-warping-based context aggregation.
Xu \etal \cite{xu2019deep} propose a multi-stage framework for video inpainting: they first use a deep flow completion network to restore the flow sequence, then perform forward and backward pixel propagation using the restored flow sequence, and finally use a pretrained image inpainting model to refine the results.
The second group uses a fixed set of frames from the entire video as reference information.
Wang \etal \cite{wang2019video} propose a two-stage model with a combination of 3D and 2D CNNs.
CPNet \cite{lee2019copy} conducts context aggregation by predicting affine matrices and applying affine transformation on fixedly sampled reference frames.

Although these video inpainting methods have shown promising results, they still suffer from ineffective usage of short-term and long-term frame reference information, and inaccurate context aggregation as discussed in Section~\ref{secIntro}.

\section{Short-Term and Long-Term Context Aggregation Network}
Given a sequence of continuous frames from a video  $X \coloneqq \{X_{1},X_{2},...,X_{T}\}$ annotated with binary masks $M \coloneqq \{M_{1},M_{2},...,M_{T}\}$, a video inpainting network outputs the restored video $\hat{Y} \coloneqq \{\hat{Y}_{1},\hat{Y}_{2},...,\hat{Y}_{T}\}$.
The goal is that $\hat{Y}$ should be spatially and temporally consistent with the ground truth video $Y \coloneqq \{Y_{1},Y_{2},...,Y_{T}\}$.

\subsection{Network Overview}
\begin{figure*}[tb]
\centering
\includegraphics[width=0.85\linewidth]{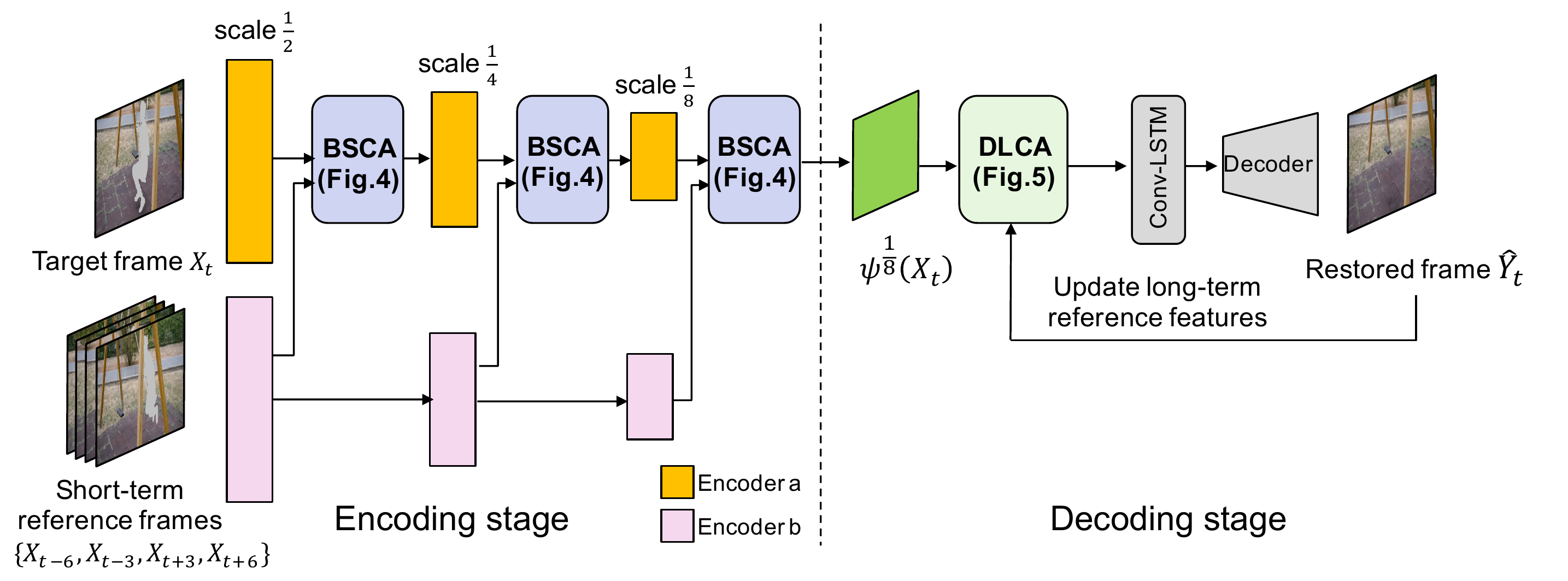}
\caption{Overview of our proposed network. In the encoding stage, we conduct Boundary-aware Short-term Context Aggregation (BSCA) (Sec.~\ref{progressive}) using short-term frame information from neighbor frames, which is beneficial to context aggregation and generating temporally consistent contents. 
In the decoding stage, we propose the Dynamic Long-term Context Aggregation (DLCA) (Sec.~\ref{dynamic}), which utilizes dynamically updated long-term frame information to refine the encoding-generated feature map.}
\label{fig:structure}

\end{figure*}

Our network is built upon a recurrent encoder-decoder architecture and processes the input video frame by frame in its temporal order.
An overview of our proposed network is illustrated in Figure \ref{fig:structure}. Different from existing methods, we start inpainting (context aggregation) at the encoding stage.
Given current target frame $X_{t}$, we choose a group of neighboring frames $\{X_{i}\}$ with $ i \in \{t-6, t-3, t+3, t+6\}$) as the short-term reference frames for $X_{t}$.
During encoding, we have two sub-encoders: encoder $a$ for the stream of target frame and encoder $b$ for the other four streams of reference frames.
The encoding process contains three feature spatial scales $\{\frac{1}{2}, \frac{1}{4}, \frac{1}{8}\}$.
At each encoding scale, we perform \textbf{Boundary-aware Short-term Context Aggregation (BSCA)} between feature maps of the target frame and those of the short-term reference frames, to fill the missing regions in the \emph{target feature map}.
This module can accurately locate and aggregate relevant bounding regions in short-term reference frames and at the same time avoid distractions caused by missing regions in the target frame.
At the decoding stage, our \textbf{Dynamic Long-term Context Aggregation (DLCA)} module refines the encoding-generated feature map using dynamically updated long-term features.
This module stores long-term frame features selected from previously \emph{restored} frames, and updates them according to their contextual correlation to the current target frame.
Intuitively, it only keeps those long-term frame features that are more contextual relevant to the current target frame.
We also adopt a convolutional LSTM (Conv-LSTM) layer to increase temporal consistency as suggested by Lai \etal \cite{lai2018learning}.
Finally, the decoder takes the refined latent feature to generate the restored frame $\hat{Y}_{t}$.
Since the missing regions are now filled with contents, we replace the target frame $X_{t}$ by the restored frame $\hat{Y}_{t}$, which provides more accurate information for the following iterations.

\begin{figure*}[tb]

\centering
\includegraphics[width=0.9\linewidth]{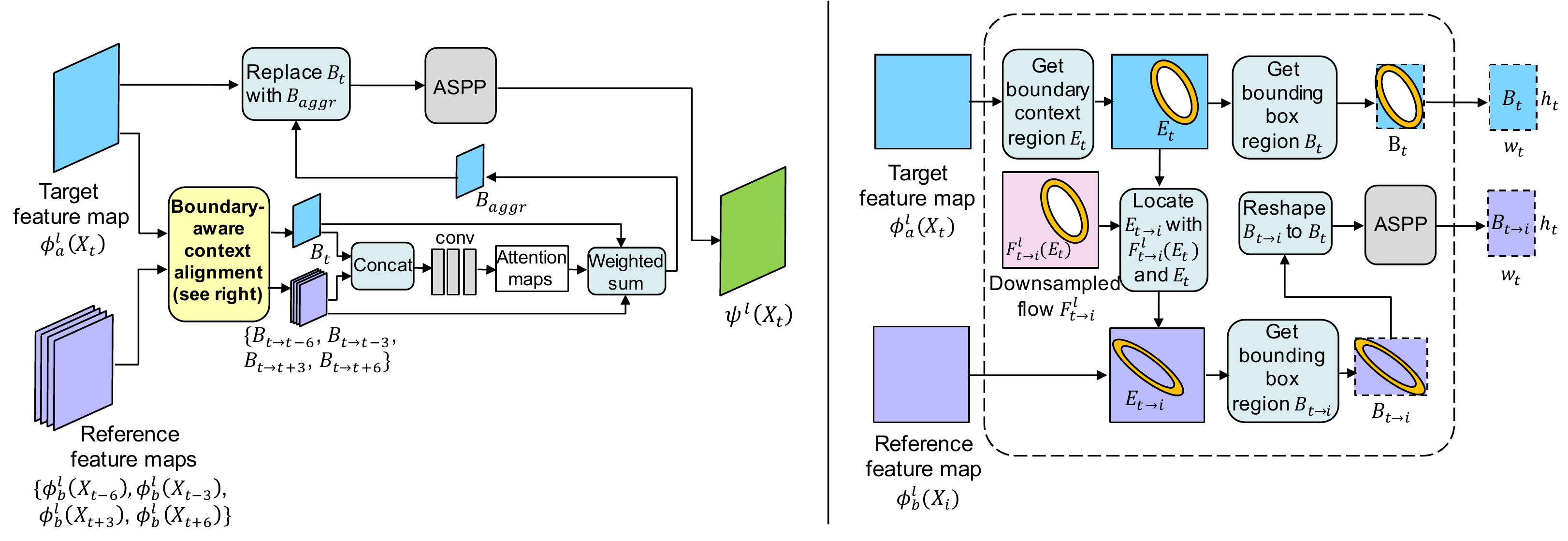}

\caption{\textbf{Left}: Boundary-aware Short-term Context Aggregation (BSCA) module. \textbf{Right}: The boundary-aware context alignment operation in BSCA. Here, $l \in \{\frac{1}{2}, \frac{1}{4}, \frac{1}{8}\}$ refers to the encoding scale.}
\label{fig:boundary}

\end{figure*}

\subsection{Boundary-aware Short-term Context Aggregation}
\label{progressive}
Optic flows between frames have been shown to be essential for alignment with short-term reference frames.
Previous optic-flow-based works \cite{kim2019deep,xu2019deep} conduct context aggregation by warping the reference frames onto the target frame.
However, missing regions in the target frame become occlusion factors and may lead to incorrect warping, as we have shown in Figure~\ref{flow}.
To alleviate this problem, we propose to utilize optic flows in a novel way: instead of using optic flows to do warping, we only use them to locate the corresponding bounding regions in the reference frame feature map that match the surrounding context of the missing regions in the target frame feature map.
Here, we define the surrounding context region as the non-missing pixels that are within a Euclidean distance $d$ ($d=8$ in our experiments) to the nearest pixels from the missing regions.

The structure of BSCA is illustrated in the left subfigure of Figure \ref{fig:boundary}.
At a certain encoding scale $l \in \{\frac{1}{2}, \frac{1}{4}, \frac{1}{8}\}$, we have the target feature map $\phi_{a}^{l}(X_{t})$ from encoder $a$ and the reference feature maps $\{\phi_{b}^{l}(X_{i})\}$ from encoder $b$ as input for boundary-aware context aggregation. We first obtain the bounding region $B_{t}$ of the missing region in $\phi_{a}^{l}(X_{t})$ and its corresponding bounding regions $\{B_{t \rightarrow i}\}$ in $\{\phi_{b}^{l}(X_{i})\}$.
Then, we apply an attention-based aggregation to combine $B_{t}$ and $\{B_{t \rightarrow i}\}$ as $B_{aggr}$.
We replace $B_{t}$ in $\phi_{a}^{l}(X_{t})$ with $B_{aggr}$ and obtain the restored target feature map $\psi^{l}(X_{t})$ , which, together with the original reference feature maps $\{\phi_{b}^{l}(X_{i})\}$, is passed on to the next encoding scale (see Figure \ref{fig:structure}).
Two essential operations in this process, \ie 1) boundary-aware context alignment and 2) attention-based aggregation, are detailed below.




\noindent\textbf{Boundary-aware Context Alignment.\:}
As illustrated in the right subfigure of Figure~\ref{fig:boundary}, the alignment operation takes the target feature map $\phi_{a}^{l}(X_{t})$ and a reference feature map $\phi_{b}^{l}(X_{i})$ as inputs.
In $\phi_{a}^{l}(X_{t})$, we denote the missing region with white color.
Then, surrounding region $E_{t}$ in $\phi_{a}^{l}(X_{t})$ is obtained (the yellow elliptical ring in $\phi_{a}^{l}(X_{t})$).
We further obtain the bounding box region of $E_{t}$ and denote it by $B_{t}$.
We use a pretrained FlowNet2 \cite{ilg2017flownet} to extract the flow information $F_{t \rightarrow i}$ between $X_{t}$ and $X_{i}$, and then we downsample $F_{t \rightarrow i}$ to $F_{t \rightarrow i}^{l}$ for current encoding scale $l$.
In $F_{t \rightarrow i}^{l}$, the corresponding flow information of $E_{t}$ is denoted as $F_{t \rightarrow i}^{l}(E_{t})$, which has the same position with $E_{t}$.
With $E_{t}$ and $F_{t \rightarrow i}^{l}(E_{t})$, we can locate the corresponding region of $E_{t}$ in $\phi_{b}^{l}(X_{i})$, which is $E_{t \rightarrow i}$ (the yellow elliptical ring in $\phi_{b}^{l}(X_{i})$).
We also obtain the bounding box region of $E_{t \rightarrow i}$ as $B_{t \rightarrow i}$, and reshape it to the shape of $B_{t}$.
To ensure the context coherence, we further refine the reshaped $B_{t \rightarrow i}$ using Atrous Spatial Pyramid Pooling (ASPP) \cite{chen2017deeplab}.
With the aligned bounding regions $B_{t}$ and $B_{t \rightarrow i}$, we can alleviate the impact of missing regions and achieve more accurate context aggregation.

\noindent\textbf{Attention-based Aggregation.\:} 
Attention-based aggregation can help find the most relevant features from the reference feature maps, and eliminate irrelevant contents, \eg newly appeared backgrounds.
We first append $B_t$ into the set $\{B_{t \rightarrow i}\}$ to get a new set $\{B_{t \rightarrow i}, B_{t}\}$, denoted as $\{B'_j\}_{j=1}^5$. Then, we concatenate the elements in the new set along the channel dimension, and apply convolutional and softmax operations across different channels to obtain the attention maps $\{A_j\}_{j=1}^5$. Finally, the attention-based aggregation is performed as follows.
\begin{equation}
B_{aggr} = \sum_{j=1}^{5}A_jB^{'}_j,
\end{equation}

We replace $B_{t}$ with the aggregated bounding region $B_{aggr}$ in the target feature map $\phi_{a}^{l}(X_{t})$.
The replaced target feature map is processed by an ASPP module to get the $\psi^{l}(X_{t})$.

\begin{figure}[tb]

\centering
\includegraphics[width=0.65\linewidth]{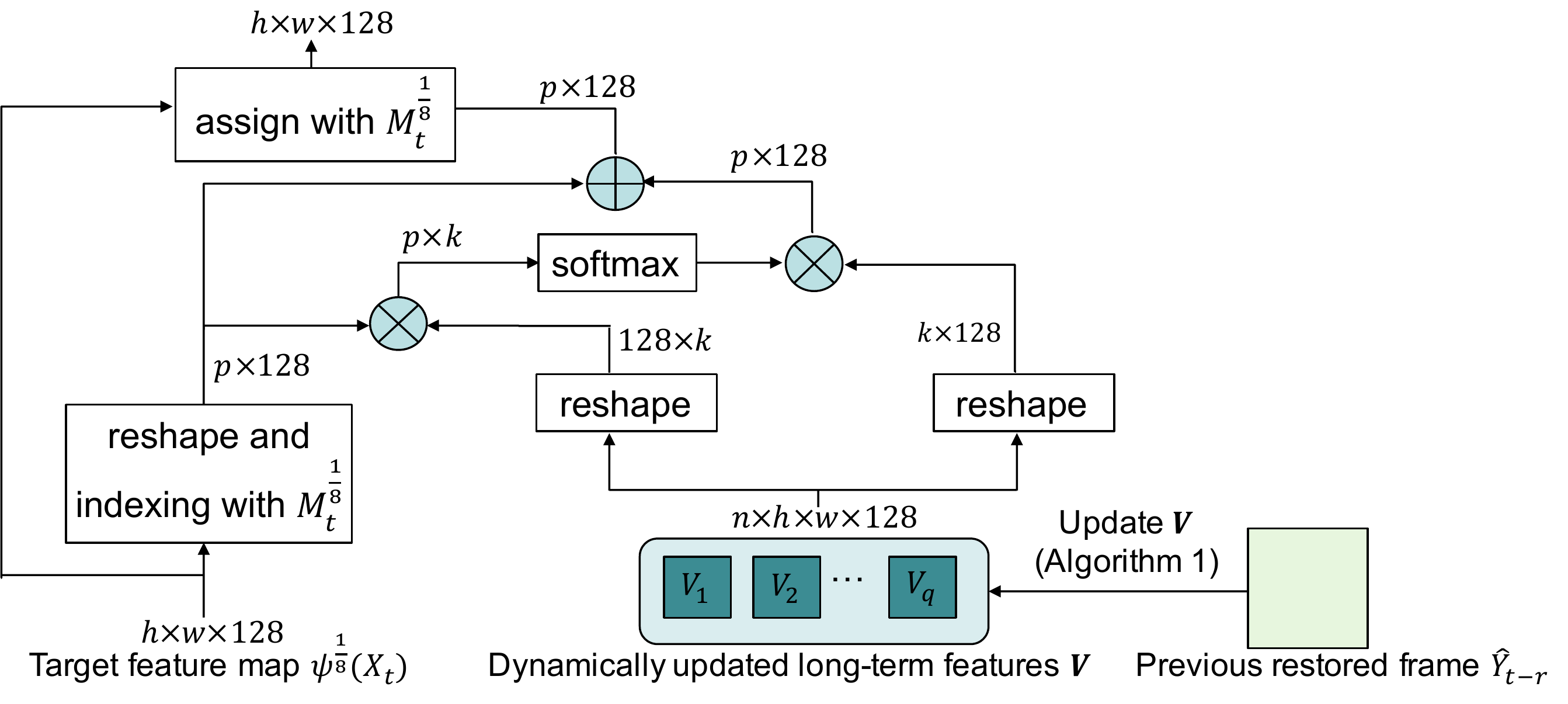}
\caption{The Dynamic Long-term Context Aggregation (DLCA) module.}
\label{fig:non_local}

\end{figure}

\subsection{Dynamic Long-term Context Aggregation}
\label{dynamic}
Fixed sampling long-term reference frames \cite{lee2019copy,oh2019onion} fail to consider the motion diversity of videos.
Thus, they may inevitably bring more irrelevant or even noisy information.
Since different videos have different motion patterns (\eg slow moving or back-and-forth moving), it results in different contextual dependency between frames.
Therefore, it is necessary that the selected long-term reference information is contextually relevant to the current target frame.
We use a dynamic strategy for the effective use of long-term reference information.
The structure of this decoding-stage context aggregation module is illustrated in Figure~\ref{fig:non_local}. 
It refines the feature map generated in the above encoding stage with 1) dynamically updated long-term features and 2) non-local-based aggregation.

\begin{algorithm}[H]
\algsetup{linenosize=\tiny}
\caption{Update Long-Term Features}
\label{ALG1}
\scriptsize
\begin{algorithmic}[1]
\REQUIRE previous restored frame $\hat{Y}_{t-r}$, current target frame $X_{t}$, long-term features $V$ 
\ENSURE updated $V$
\STATE $\text{distance} = [\;]$
\STATE $U_{X_{t}} = \text{Encoder\_b}(X_{t})$
\STATE $U_{\hat{Y}_{t-r}} = \text{Encoder\_b}(\hat{Y}_{t-r})$
\STATE $d_{\hat{Y}_{t-r}} = \|U_{X_{t}} - U_{\hat{Y}_{t-r}}\|_{1}$
\FOR{$V_{r}$ in $V$} 
\STATE $d_{r} = \|U_{X_{t}} - V_{r}\|_{1}$
\STATE $\text{distance.append}(d_{r})$
\ENDFOR
\STATE $d_{max}, max = \text{get\_max\_and\_index}(\text{distance})$ 
\IF{$d_{\hat{Y}_{t-r}} < d_{max}$}
\STATE $\text{$V$.remove}(V_{max})$
\STATE $\text{$V$.append}(U_{\hat{Y}_{t-r}})$
\ENDIF

\end{algorithmic}
\end{algorithm}

\noindent\textbf{Dynamically Updated Long-Term Features.\:}
DLCA stores the features of the previously \emph{restored} frames that are most relevant (in feature space) to the current target frame.
Specifically,  $V \coloneqq \{V_{1}, V_{2}, ..., V_{q}\}$ stores a set of long-term feature maps with the length $q$, which are updated dynamically following Algorithm \ref{ALG1}.
At each inpainting iteration, it checks whether the feature map of a long-term frame $\hat{Y}_{t-r}$ ($r$ is the parameter that defines how far from the current target frame to look back) can be incorporated into the $V$ set according to its $L_{1}$ distance to target frame $X_{t}$ in the feature space. Let $U_{X_{t}}$ and $U_{\hat{Y}_{t-r}}$ be the feature maps of $X_{t}$ and $\hat{Y}_{t-r}$ respectively, if the $L_{1}$ distance between $U_{\hat{Y}_{t-r}}$ and $U_{X_{t}}$ is smaller than the maximum distance between a feature map in the current $V$ set to $U_{X_{t}}$, then $U_{\hat{Y}_{t-r}}$ will replace the corresponding feature map (that has the maximum distance) in the $V$ set. Note that these feature maps can be obtained using our encoder $b$.
We suggest $r \geq 7$ to exploit long-term information (as short-term information from $\hat{Y}_{t-6/t-3}$ has already been considered by our BSCA module). 
At the beginning when $t<r$, we simply set $r=|t-r|$ to use all restored frames so far.
With this dynamic updating policy, DLCA can automatically adjust the stored long-term frame features and remove irrelevant ones, regarding each target frame. 

\noindent\textbf{Non-local-based Aggregation.\:}
Based on the long-term feature set $V$, we follow a typical approach \cite{wang2018non} to perform non-local-based context aggregation between the target feature map $\psi^{\frac{1}{8}}(X_{t})$ and feature maps stored in $V$, as shown in Figure~\ref{fig:non_local}.
Softmax is applied to obtain the normalized soft attention map over feature maps in $V$.
The attention map is then utilized as weights to compute an aggregated feature map from $V$ via weighted summation.
Finally, the aggregated feature map replaces the feature map of missing regions.


\subsection{Loss Function}
The loss function used for training is:

\begin{equation}
\mathcal{L}_{total} = \mathcal{L}_{\text{rec}} + \lambda_{\text{mre}}\mathcal{L}_{\text{mre}} + \lambda_{\text{per}}\mathcal{L}_{\text{per}} + \lambda_{\text{style}}\mathcal{L}_{\text{style}},
\end{equation}
Here, $\mathcal{L}_{\text{rec}}$, $\mathcal{L}_{\text{mre}}$, $\mathcal{L}_{\text{per}}$, and $\mathcal{L}_{\text{style}}$ denote reconstruction loss, reconstruction loss of mask region, perceptual loss, and style loss respectively.
The balancing weights $\lambda_{\text{mre}}$, $\lambda_{\text{per}}$ and $\lambda_{\text{style}}$ are empirically set to 2, 0.01, and 1, respectively. 

The reconstruction loss and the reconstruction loss of the mask region are defined on pixels:

\begin{equation}
\mathcal{L}_{\text{rec}} = \sum_{t}^{T}\|\hat{Y}_{t} - Y_{t}\|_{1}.
\end{equation}

\begin{equation}
\mathcal{L}_{\text{mre}} = \sum_{t}^{T}\|(1 - M_{t})\odot(\hat{Y}_{t} - Y_{t})\|_{1},
\end{equation}
where $\odot$ is the element-wise multiplication.
To further enhance inpainting quality, we include two additional loss functions: perceptual loss \cite{johnson2016perceptual} and  style loss,

\begin{equation}
\mathcal{L}_{\text{per}} = \sum_{t}^{T}\sum_{s}^{S}\frac{\|\sigma_{s}(\hat{Y}_{t}) -\sigma_{s}( Y_{t})\|_{1}}{S},
\end{equation}

\begin{equation}
\mathcal{L}_{\text{style}} = \sum_{t}^{T}\sum_{s}^{S}\frac{\|G_{s}^{\sigma}(\hat{Y}_{t}) -G_{s}^{\sigma}( Y_{t})\|_{1}}{S},
\end{equation}
where $\sigma_{s}$ is the $s$-th layer output of an ImageNet-pretrained VGG-16 \cite{simonyan2014very} network, $S$ is the number of chosen layers (\ie $relu_{2\_2}$, $relu_{3\_3}$ and $relu_{4\_3}$), and
$G$ denotes the gram matrix multiplication \cite{gatys2016image}.
\section{Experiments}
We evaluate and compare our model with state-of-the-art models qualitatively and quantitatively. We also conduct a comprehensive ablation study on our proposed model.

\begin{table*}[thb]
\renewcommand\arraystretch{1.05}

\caption{Quantitative comparisons on YouTube-VOS and DAVIS datasets under three mask settings regarding 3 performance metrics: PSNR (higher is better), SSIM (higher is better) and VFID (lower is better). The rightmost column shows the average execution time to inpaint one video. The best results are in \textbf{bold}.}
\begin{center}
\scalebox{0.7}
{
\begin{tabular}{l|ccc|ccc|ccc|c}
      			
      			\hline
      			\multicolumn{11}{c}{YouTube-VOS} \\
      			\hline
      	 \multirow{2}{*}{Model}  & \multicolumn{3}{c|}{Square mask} & \multicolumn{3}{c|}{Irregular mask} & \multicolumn{3}{c|}{Object mask} & \multirow{2}{*}{Time (sec.)} \\ 
     			& PSNR & SSIM & VFID  & PSNR & SSIM & VFID & PSNR & SSIM & VFID \\ \hline 
		VINet \cite{kim2019deep} & 26.92 & 0.843 & 0.103 & 27.33  & 0.848 & 0.082 & 26.61 & 0.838 & 0.118 &  \textbf{33.6} \\	 
	CPNet \cite{lee2019copy} & 27.24 & 0.847 & 0.087 & 27.50 & 0.852 & 0.051 & 27.02 &  0.845 & 0.087 & 48.5 \\  
		FGNet \cite{xu2019deep}	 & 27.71 & 0.856 & 0.082 & 27.91  & 0.859 & 0.056 & 27.32 & 0.849 & 0.083 &  276.3 \\ \hline
      		\textbf{Ours} & \textbf{27.76} & \textbf{0.858} & \textbf{0.076} &  \textbf{28.12}  & \textbf{0.866} & \textbf{0.047} & \textbf{27.45} & \textbf{0.853} & \textbf{0.075} & 35.4 \\ 
      		\hline
      		\hline
      			\multicolumn{11}{c}{DAVIS} \\
      			\hline
      	 \multirow{2}{*}{Model}  & \multicolumn{3}{c|}{Square mask} & \multicolumn{3}{c|}{Irregular mask} & \multicolumn{3}{c|}{Object mask} & \multirow{2}{*}{Time (sec.)} \\ 
     			& PSNR & SSIM & VFID  & PSNR & SSIM & VFID & PSNR & SSIM & VFID \\ \hline 
		VINet \cite{kim2019deep} & 27.88 & 0.863 & 0.060 & 28.67  & 0.874 & 0.043 & 27.02 & 0.850 & 0.068 &  \textbf{19.7} \\	 
	CPNet \cite{lee2019copy}	 & 27.92 & 0.862 & 0.049 & 28.81 & 0.876 & 0.031 & 27.48 &  0.855 & 0.049 & 28.2 \\  
		FGNet \cite{xu2019deep}	 & 28.32 & 0.870 & 0.045 & 29.37  & 0.880 & 0.033 & \textbf{28.18} & 0.864 & 0.046 & 194.8 \\ \hline
      		\textbf{Ours} & \textbf{28.50} & \textbf{0.872} & \textbf{0.038} &  \textbf{29.56}  & \textbf{0.883} & \textbf{0.027} & 28.13 & \textbf{0.867} & \textbf{0.042}  & 21.5 \\ 
      		\hline
\end{tabular}    
}
\end{center}

\label{tab0}
\end{table*}

\noindent\textbf{Datasets.\:}
Following previous works \cite{kim2019deep,lee2019copy}, we train and evaluate our model on YouTube-VOS \cite{xu2018youtube} and DAVIS \cite{perazzi2016benchmark} datasets.
For YouTube-VOS, we use the 3471 training videos for training, and the 508 test videos for testing.
For DAVIS, we use the 60 unlabeled videos to fine tune a pretrained model on YouTube-VOS, and the 90 videos with object mask annotations for testing.
All video frames are resized to $424 \times 240$ , and no pre-processing or post-processing is applied.

\noindent\textbf{Mask Settings.\:}
To simulate the diverse and ever-changing real-world scenarios, we consider the following three mask settings for training and testing.
\begin{itemize}
\item Square mask: The same square region for all frames in a video, but has a random location and a random size ranging from $40 \times 40$ to $160 \times 160$ for different videos.

\item Irregular mask: We use the irregular mask dataset \cite{liu2018image} that consists of masks with arbitrary shapes and random locations.

\item Object mask: Following \cite{kim2019deep,xu2019deep}, we use the foreground object masks in DAVIS \cite{perazzi2016benchmark} which has continuous motion and realistic appearance.
Note that when quantitatively testing object masks on DAVIS dataset, we shuffle its video-mask pairs for more reasonable result, as it is a dataset for object removal and the ground-truth background (after removal) is unknown (without shuffling, the original objects will become the ground truth). 
\end{itemize}

\noindent\textbf{Baseline Models.\:}
We compare our model with three state-of-the-art video inpainting models: 1) VINet \cite{kim2019deep}, a recurrent encoder-decoder network with flow-warping-based context aggregation; 2) CPNet \cite{lee2019copy}, which conducts context aggregation by predicting affine matrices and applying affine transformation on fixedly sampled reference frames; and 3) FGNet \cite{xu2019deep}, which consists of three stages: first restores flow between frames, then performs forward and backward warping with the restored flow, and finally utilizes an image inpainting model for post-processing.


\begin{figure*}[t!]

    \centering
    \setlength{\tabcolsep}{0.10em}
    {

\begin{tabular}{ccccccc}
        
\rotatebox[origin=l]{90}{\scriptsize Input}&
\includegraphics[width=1.9cm]{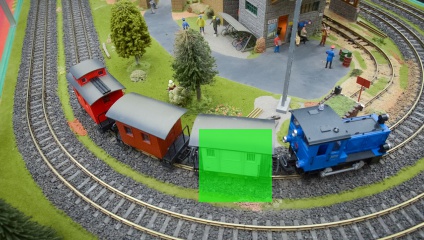}&
\includegraphics[width=1.9cm]{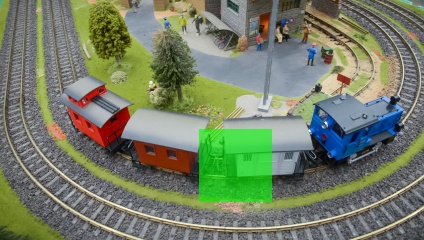}&
\includegraphics[width=1.9cm]{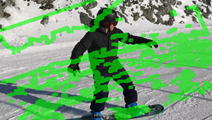}&
\includegraphics[width=1.9cm]{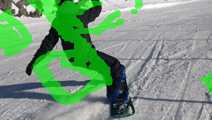}&
\includegraphics[width=1.9cm]{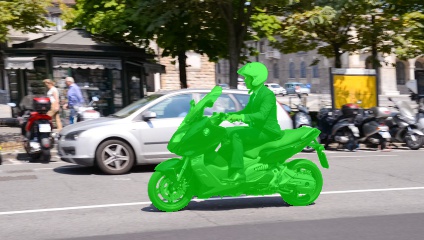}&
\includegraphics[width=1.9cm]{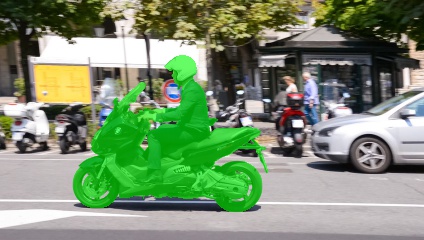}\\

\addlinespace[-0.15em]

\rotatebox[origin=l]{90}{\scriptsize VINet}&
\includegraphics[width=1.9cm]{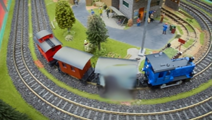}&
\includegraphics[width=1.9cm]{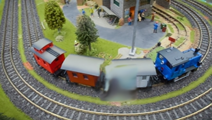}&
\includegraphics[width=1.9cm]{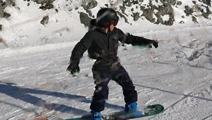}&
\includegraphics[width=1.9cm]{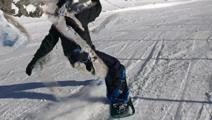}&
\includegraphics[width=1.9cm]{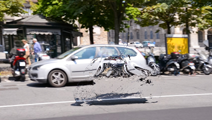}&
\includegraphics[width=1.9cm]{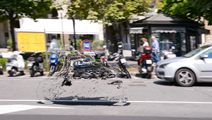}\\

\addlinespace[-0.15em]

\rotatebox[origin=l]{90}{\scriptsize CPNet}&
\includegraphics[width=1.9cm]{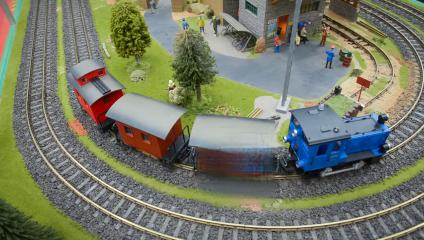}&
\includegraphics[width=1.9cm]{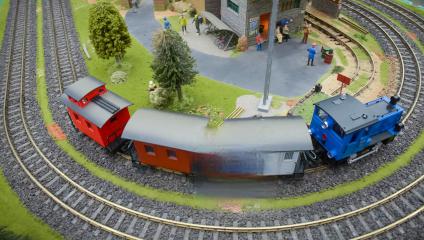}&
\includegraphics[width=1.9cm]{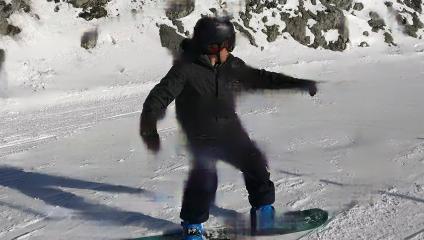}&
\includegraphics[width=1.9cm]{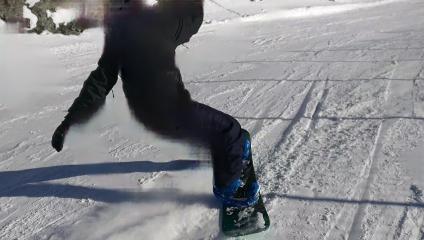}&
\includegraphics[width=1.9cm]{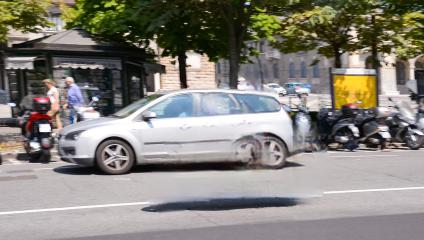}&
\includegraphics[width=1.9cm]{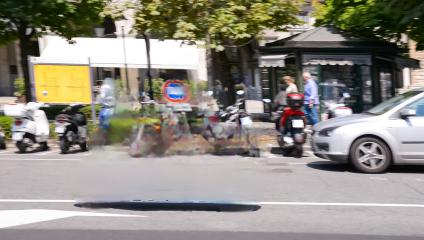}\\

\addlinespace[-0.15em]

\rotatebox[origin=l]{90}{\scriptsize FGNet}&
\includegraphics[width=1.9cm]{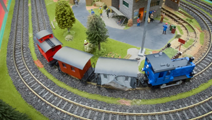}&
\includegraphics[width=1.9cm]{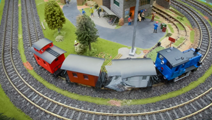}&
\includegraphics[width=1.9cm]{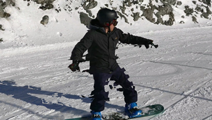}&
\includegraphics[width=1.9cm]{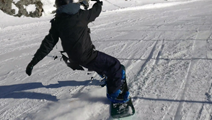}&
\includegraphics[width=1.9cm]{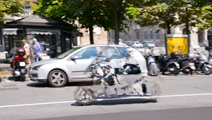}&
\includegraphics[width=1.9cm]{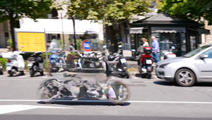}\\

\addlinespace[-0.15em]

\rotatebox[origin=l]{90}{\scriptsize Ours}&
\includegraphics[width=1.9cm]{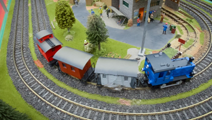}&
\includegraphics[width=1.9cm]{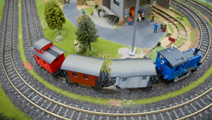}&
\includegraphics[width=1.9cm]{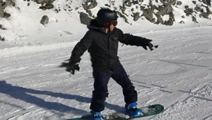}&
\includegraphics[width=1.9cm]{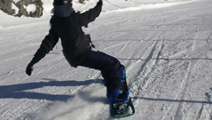}&
\includegraphics[width=1.9cm]{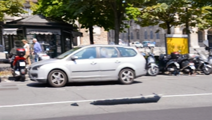}&
\includegraphics[width=1.9cm]{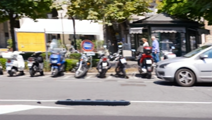}\\

\end{tabular}  

}

\caption{Qualitative comparison of our proposed model with baseline models on DAVIS dataset. Better viewed at zoom level 400\%. More video results can be found in supplementary material. }
\label{QualitativeCompare}

\end{figure*}

\subsection{Quantitative Results}
We consider three metrics for evaluation: 1) \textit{Peak Signal-to-Noise Ratio} (PSNR, measures image distortion), 2) \textit{Structural Similarity} (SSIM, measures structure similarity) and 3) the video-based Fréchet Inception Distance (VFID, a video perceptual measure known to match well with human perception) \cite{wang2018video}.
As shown in Table~\ref{tab0}, our model outperforms all baseline models according to all three metrics across all three mask settings on both datasets, a clear advantage of using both short-term and long-term information.
In terms of execution time, our model is comparable to VINet, which has the least average execution time but worse performance than all other three models.
Overall, our model achieves the best trade-off between performance and execution time on the two test sets.

\subsection{Qualitative Results}
To further inspect the visual quality of the inpainted videos, we show, in Figure~\ref{QualitativeCompare}, three examples of the inpainted frames by our model and the compared baselines. 
As can be observed, frames inpainted by our models are generally of much higher quality than those by VINet or CPNet, and also perceptibly better than the state-of-the-art model FGNet.
For example, in the third example (right two columns), the car structures generated by VINet are highly distorted. This is mainly caused by the occlusion effect of mask regions in the target frame, and its limited exploration of long-term information.
CPNet was able to restore the rough structures of the car with more information from its fixedly sampled long-term reference frames.
However, blurriness or overlapping can still be found since those fixed-term reference frames also bring in a significant amount of irrelevant contexts.
FGNet in general achieves sharper results than VINet or CPNet.
However, it also generates artifacts in this example.
This can be ascribed to inaccurate flow inpainting in the first stage of FGNet.
In contrast, our model can generate more plausible contents with high spatial and temporal consistency.

\subsection{User Study}
We also conduct a user study to verify the effectiveness of our proposed network.
We recruited 50 volunteers for this user study.
We randomly select 20 videos from DAVIS test set.
For each video, the original video with object mask and the anonymized results from VINet, CPNet, FGNet and our model are presented to the volunteers.
The volunteers are then asked to rank the four models with 1, 2, 3 and 4 (1 is the best and 4 is the worst) based on the perceptual quality of the inpainted videos.
The result in terms of the percentage of rank scores received by different models is shown in Figure~\ref{user}.
Our model receives significantly more votes for rank 1 (the best) than the other three models, which verifies that our model can indeed generate more plausible results than existing models.

\pgfplotstableread[col sep=comma,header=false]{
rank 1,54,26,16,4
rank 2,24,30,36,10
rank 3,18,34,28,20
rank 4,4,10,20,66
}\data

\pgfplotsset{
percentage plot/.style={
    point meta=explicit,
nodes near coords align=vertical,
    yticklabel=\pgfmathprintnumber{\tick}\,$\%$,
    ymin=0,
    ymax=100,
    legend style={at={(0.5,-0.4)}, anchor=north,legend columns=-1},
    enlarge y limits={upper,value=0},
visualization depends on={y \as \originalvalue}
},
percentage series/.style={
    table/y expr=\thisrow{#1},table/meta=#1
}
}

\begin{figure}
\begin{center}

\begin{tikzpicture}
\begin{axis}[
axis on top,
width=10cm,
height=3cm,
ylabel=percentage,
xlabel=\empty,
percentage plot,
ybar=0pt,
bar width=0.3cm,
enlarge x limits=0.25,
symbolic x coords={rank 1, rank 2, rank 3, rank 4},
xtick=data
]
\addplot table [percentage series=1] {\data};
\addplot table [percentage series=2] {\data};
\addplot table [percentage series=3] {\data};
\addplot table [percentage series=4] {\data};
\legend{Ours,FGNet,CPNet,VINET}
\end{axis}
\end{tikzpicture}
    
    \caption{User study results. For each rank (1 is the best), we collected 1000 votes (20 videos * 50 volunteers) in total. The y-axis indicates, within each rank, the percentage (out of 1000 notes) of the votes received by different models.}
    
    \label{user}
\end{center}
    
\end{figure}
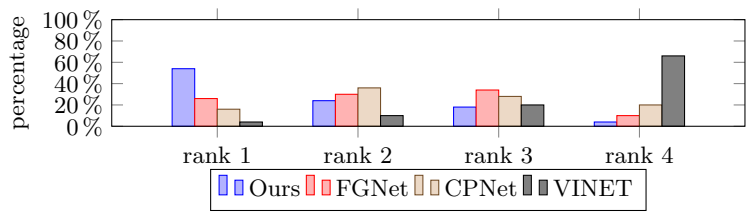

\subsection{Ablation Study}
We investigate the effectiveness of the two components of our network: Boundary-aware Short-term Context Aggregation (BSCA) and Dynamic Long-term Context Aggregation (DLCA).
In Table~\ref{tab1}, we report all the quantitative results under different ablation settings on the DAVIS test set with shuffled object masks.

\begin{figure}[t!]
\centering
\setlength{\tabcolsep}{0.1em}
{
\begin{tabular}{cccc}

\includegraphics[width=2.7cm]{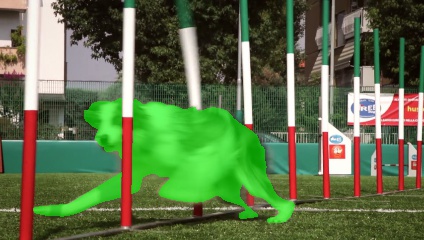}&
\includegraphics[width=2.7cm]{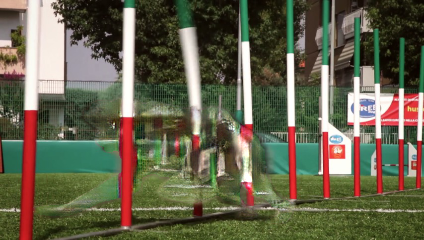}&
\includegraphics[width=2.7cm]{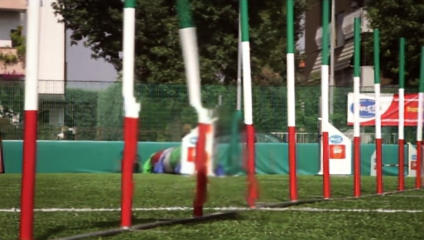}&
\includegraphics[width=2.7cm]{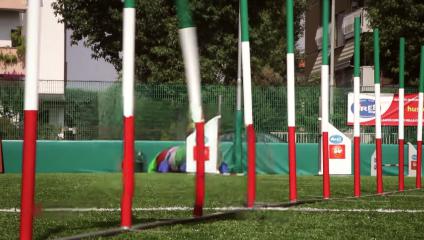}

\\

 \scriptsize (a) Input & \scriptsize (b) Ours (w/o BSCA) & \scriptsize (c) Ours (scale $\frac{1}{8}$)  & \scriptsize (d) Ours

  \\

\end{tabular}
}

\caption{Ablation study on BSCA. Better viewed at zoom level 400\%.}

\label{ablation1} 
\end{figure}

\noindent\textbf{Effectiveness of BSCA.\:}
As we described in Sec.~\ref{progressive}, the purpose of BSCA is to alleviate the negative effects of missing regions in the target frame. Table~\ref{tab1} compares our full model with its two variants: 1) ``w/o BSCA" (the first row in Table~\ref{tab1}), which removes the BSCA module and directly uses the flows to warp reference feature maps onto the target feature map as \cite{kim2019deep}.
Then it concatenates the warped reference features with the target feature map as the input for attention-based aggregation; 2) ``scale $\frac{1}{8}$" (the second row in Table~\ref{tab1}), which performs the BSCA module only at the $\frac{1}{8}$ encoding scale.
The performance drop of these two variants justifies the effectiveness of the BSCA module and the multi-scale design at the encoding stage.
As we further show in Figure~\ref{ablation1}, the model without BSCA suffers from inaccurate feature alignment due to the occlusion effect of missing regions in the flows, thus producing distorted contents.
Using BSCA only at the $\frac{1}{8}$ encoding scale apparently improves the results, but is still affected by the distortions from the previous scales.
In contrast, using BSCA at multiple encoding scales lead to better results with temporally consistent details.

\begin{table}[thb]
\renewcommand\arraystretch{1.0}

\caption{Comparisons of different settings on BSCA and DLCA.}
\begin{center}
\scalebox{0.75}
{
\begin{tabular}{c|c|c|c|ccc}
      			\hline
      
     BSCA (scale $\frac{1}{8}$) & BSCA  & DLCA (fixed) & DLCA	& PSNR & SSIM & VFID   \\ \hline
     & &  & \checkmark			 & 27.17 & 0.847 & 0.063 \\ \hline
		  \checkmark & & & \checkmark  & 27.72 & 0.859 & 0.052  \\	 \hline
		  & \checkmark  & &  & 27.58 & 0.858 & 0.056 \\ \hline
		 & \checkmark &  \checkmark &  & 27.85 & 0.862 & 0.048  \\ \hline
      		  & \checkmark &  & \checkmark & \textbf{28.13} & \textbf{0.867} & \textbf{0.042} \\ \hline 
\end{tabular}    
}
\end{center}

\label{tab1}
\end{table}

\begin{figure}[t!]
\centering
\setlength{\tabcolsep}{0.1em}
{
\begin{tabular}{cccccc}

\includegraphics[width=1.95cm]{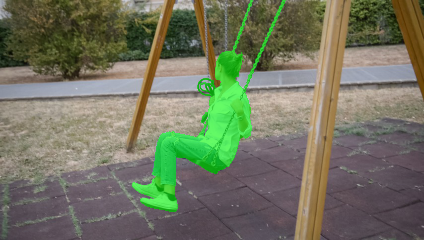}&
\includegraphics[width=1.95cm]{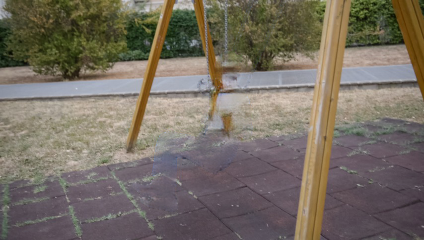}&
\includegraphics[width=1.95cm]{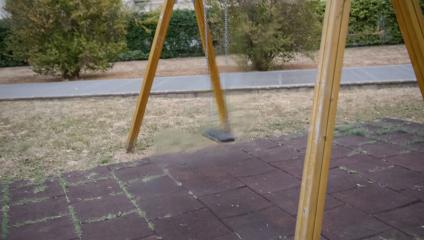}&
\includegraphics[width=1.95cm]{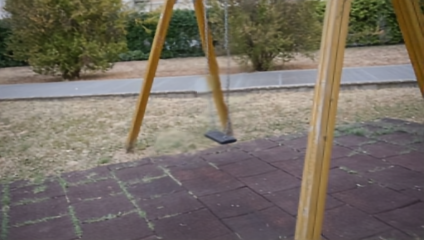}&
\includegraphics[width=1.95cm]{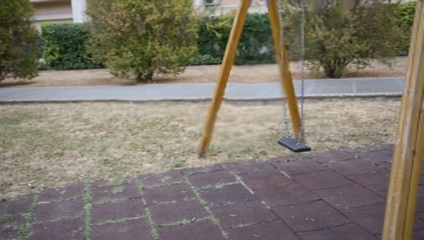}&
\includegraphics[width=1.95cm]{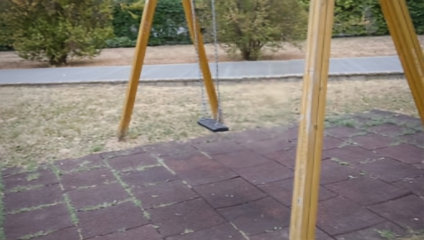}
\\

\scriptsize (a) Input  & \scriptsize (b) Ours  & \scriptsize (c) Ours  & \scriptsize (d) Ours & \scriptsize (e) $\hat{Y}_{t-13}$  & \scriptsize (f) $\hat{Y}_{t-25}$ 
  \\
  
  \scriptsize  (at time $t$) & \scriptsize  (w/o DLCA) & \scriptsize  (Fixed) & \scriptsize  & \scriptsize  (not in $V$)  & \scriptsize  (in $V$)
  \\

\end{tabular}
}

\caption{Ablation study on DLCA. Better viewed at zoom level 400\%.}

\label{ablation2} 
\end{figure}

\noindent\textbf{Effectiveness of DLCA.\:}
We test two other variants of our full model regarding the DLCA module: 1) ``w/o DLCA" (the third row in Table~\ref{tab1}), which directly removes the DLCA module; and 2) ``fixed" (the fourth row in Table~\ref{tab1}), which keeps the DLCA module but uses fixedly sampled reference features (rather than dynamic updated ones) that takes one frame for every five frames out of the entire input video sequence.
Both variants exhibit performance degradation. 
Although fixedly sampled reference features can help, it is still less effective than using our dynamically updated long-term features.
As shown in Figure~\ref{ablation2}, the model without DLCA module (w/o DLCA) fails to recover the background after object removal due to the lack of long-term frame information.
Although the model with fixedly sampled reference features successfully restores the background, blurriness and artifacts can be still be found.
Figure~\ref{ablation2} (e) and Figure~\ref{ablation2} (f) further illustrate the dynamic characteristic of our dynamic updating rule, which can effectively avoid irrelevant reference frames (\eg $\hat{Y}_{t-13}$ is not in our long-term feature set $V$) for the current target frame.

\noindent\textbf{Dynamically Updated Long-term Features.\:}
We investigate the impact of different lengths of dynamically updated long-term features $V$ on the inpainting results.
Small lengths of $q$ is insufficient to capture long-term frame information, resulting in inferior performance. 
On the contrary, large lengths will include more irrelevant reference frames, which also leads to degraded performance.
The best result is achieved at length $q=10$.
For the parameter $r$ (long-term range), we empirically find that $r=9$ works well across different settings.

\section{Conclusion}
We studied the problem of video inpainting and addressed three limitations of existing methods: 1) ineffective usage of short-term or long-term reference frames; 2) inaccurate short-term context aggregation caused by missing regions in the target frame; and 3) fixed sampling of long-term contextual information. 
We therefore proposed a Short-term, and Long-term Context Aggregation Network with two complementary modules for the effective exploitation of both short-term and long-term information. 
We have empirically demonstrated the effectiveness of our proposed approach on benchmark datasets and provided a comprehensive understanding of each module of our model. 

\section*{Acknowledgement}
This research was supported by Australian Research Council Projects FL-170100117, IH-180100002, IC-190100031, LE-200100049.

%
%
\bibliographystyle{splncs04}
\bibliography{main}
\end{document}